\documentclass{ceurart}

\usepackage{tabularx}

\newtheorem{example}{Example}

\newtheorem{definition}{Definition}

\newtheorem{notation}{Notation}

\newcommand{\tdn}{\ensuremath{\mathit{N}}}

\newcommand{\sources}{\ensuremath{\mathcal{S}}}
\newcommand{\objects}{\ensuremath{\mathcal{O}}}

\newcommand{\reports}{\ensuremath{\mathcal{P}}}
\newcommand{\val}{\ensuremath{\operatorname{val}}}

\newcommand{\args}{\ensuremath{\mathcal{A}}}
\newcommand{\attacker}{\ensuremath{\mathcal{R}^{-}}}
\newcommand{\supporter}{\ensuremath{\mathcal{R}^{+}}}

\newcommand{\domain}{\ensuremath{\mathcal{D}}}
\newcommand{\baseScore}{\ensuremath{\tau}}

\newcommand{\obj}{\operatorname{obj}}

\begin{document}

\copyrightyear{2021}
\copyrightclause{Copyright for this paper by its authors.
  Use permitted under Creative Commons License Attribution 4.0
  International (CC BY 4.0).}

\conference{2nd International Workshop on Argumentation for eXplainable AI (ArgXAI), 16 September 2024.}

\title{Applying Attribution Explanations in Truth-Discovery Quantitative Bipolar Argumentation Frameworks}

\author[1]{Xiang Yin}[%
orcid=000-0002-6096-9943,
email=x.yin20@imperial.ac.uk
]
\address[1]{Imperial College London, UK}

\author[2]{Nico Potyka}[%
orcid=0000-0003-1749-5233,
email=potykan@cardiff.ac.uk
]
\address[2]{Cardiff University, UK}

\author[1]{Francesca Toni}[%
orcid=0000-0001-8194-1459,
email=ft@imperial.ac.uk
]


\begin{abstract}
Explaining the strength of arguments under gradual semantics is receiving increasing attention.
For example, various studies in the literature offer explanations by computing the attribution scores of arguments or edges in Quantitative Bipolar Argumentation Frameworks (QBAFs). These explanations, known as \emph{Argument Attribution Explanations (AAEs)} and \emph{Relation Attribution Explanations (RAEs)}, commonly employ \emph{removal-based} and \emph{Shapley-based} techniques for computing the attribution scores.
While AAEs and RAEs have proven useful in several applications with acyclic QBAFs, they remain largely unexplored for cyclic QBAFs.
Furthermore, existing applications tend to focus solely on either AAEs or RAEs, but do not compare them directly.
In this paper, we apply both AAEs and RAEs, to Truth Discovery QBAFs (TD-QBAFs), which assess the trustworthiness of sources (e.g., websites) and their claims (e.g., the severity of a virus), and feature complex cycles. We find that both AAEs and RAEs can provide interesting explanations and can 
give non-trivial and surprising insights.
\end{abstract}


\maketitle

\section{Introduction}

\begin{figure}[t]
    \centering
    \includegraphics[width=0.8\columnwidth]{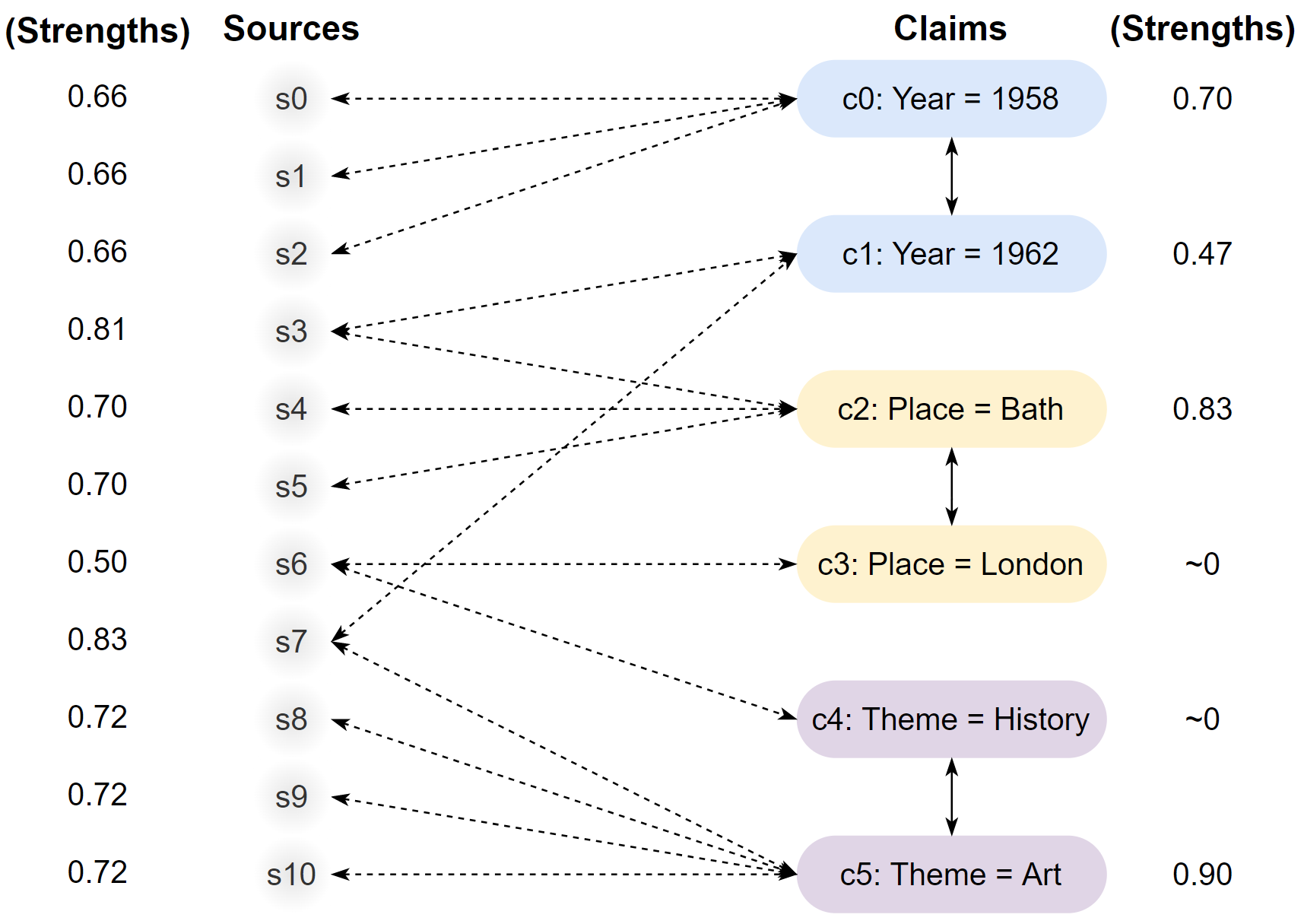}
    \caption{Example of a TD-QBAF. (Nodes are arguments, where the $s_i$ and $c_i$ are identifiers for the \emph{source} and \emph{claim} arguments, respectively (for ease of reference). Solid and dashed edges indicate \emph{attack} and \emph{support}, respectively.)}
    \label{fig_tdqbaf}
\end{figure}

Abstract argumentation Frameworks (AFs)~\cite{Dung_95} are promising tools in the Explainable AI (XAI) field~\cite{mittelstadt2019explaining} due to their transparency and interpretability, as well as their ability to support reasoning about conflicting information~\cite{vcyras2021argumentative,potyka2021interpreting,potyka2023explaining}.
Quantitative Bipolar AFs (QBAFs)~\cite{baroni2015automatic} are an extension of traditional AFs, which consider the \emph{(dialectical) strength} of arguments and the \emph{support} relation between arguments.
In QBAFs, each argument has a \emph{base score}, and its dialectical strength is computed by \emph{gradual semantics} based on its base score and the strength of its attackers and supporters~\cite{baroni2019fine}.
QBAFs can be deployed to support several applications like product recommendation \cite{rago2018argumentation}, review aggregation \cite{cocarascu2019extracting} or stance aggregation \cite{kotonya2019gradual}.

Another interesting application that has been considered recently are truth discovery networks \cite{singleton2022towards,singleton2020link,PotykaBooth24}.
Figure~\ref{fig_tdqbaf} shows an example of a \emph{Truth-Discovery QBAFs (TD-QBAF)} to evaluate the trustworthiness of sources and the reliability of claims made about an exhibition. We have $11$ sources and $6$ claims, each represented as an abstract argument. The nodes on the left represent the $11$ source arguments ($s0$ to $s10$), while the ones on the right represents the $6$ claim arguments. 
The claim arguments are categorized into three types — year, place, and theme of the exhibition — each distinguished by different colors. 
For pairs of contradictory claims, where different values are asserted for the same object, a bi-directional attack relationship is introduced between the claims. For each report (one for each pair of source and claim), a bi-directional support relationship is established between the source and the claim.
Following \cite{PotykaBooth24}, we use a base score of $0.5$ for source argument (we are initially indifferent about the trustworthiness of a source), and a base score of $0$ for claims (we do not believe claims without evidence). We compute the dialectical strength of arguments using the \emph{Quadratic Energy (QE)} gradual semantics~\cite{Potyka18}, and the final strengths of arguments are displayed on their side in Figure \ref{fig_tdqbaf}.
While the strength values seem plausible, it can be challenging to understand why certain claims and sources receive higher or lower trust scores. 

To address this problem, attribution explanations (AEs) have been proposed. Specifically, given an argument of interest (\emph{topic argument}) in a QBAF, AEs can explain the impact of arguments on the topic argument.
AEs can be broadly categorized into \emph{Argument Attribution Explanations (AAEs)} (e.g., ~\cite{delobelle2019interpretability,vcyras2022dispute,AAE_ECAI}) and \emph{Relation Attribution Explanations (RAEs)} (e.g., ~\cite{amgoud2017measuring,YIN_RAE_IJCAI}).
AAEs explain the strength of the topic argument by assigning \emph{attribution scores} to arguments: the greater the attribution score, the greater the argument's contribution to the topic argument. Similarly, RAEs assign the attribution scores to edges to measure their contribution. \emph{Removal-based} and \emph{Shapley-based} techniques are commonly used for computing the attribution scores.

However, most existing studies focus on explaining acyclic QBAFs rather than cyclic ones, leaving a gap in understanding the complexities of the latter. In addition, current research typically examines only one type of attribution — either AAEs or RAEs — without providing a comprehensive comparison of both methods.
In this paper, we aim to address these gaps by investigating the applicability of removal and Shapley-based AAEs and RAEs in the context of cyclic TD-QBAFs. Furthermore, we offer a comprehensive comparison between them to better understand the applicability of these AEs.



\section{Preliminaries}

\subsection{QBAFs and the QE Gradual Semantics}
We briefly recall the definition of QBAFs and the QE gradual semantics~\cite{Potyka18}.
\begin{definition}[QBAF]
\label{def_qbaf}
A \emph{Quantitative Bipolar Argumentation Framework (QBAF)} is a quadruple $\mathcal{Q}=\langle \mathcal{A}, \mathcal{R}^{-}, \mathcal{R}^{+}, \tau \rangle$ consisting of a finite set of \emph{arguments} $\mathcal{A}$, binary relations of \emph{attack} $\mathcal{R}^{-} \subseteq \mathcal{A} \times \mathcal{A}$ and \emph{support} $\mathcal{R}^{+} \subseteq \mathcal{A} \times \mathcal{A}$ $(\mathcal{R}^{-} \cap \mathcal{R}^{+} = \emptyset)$ and a \emph{base score function} $\tau:\mathcal{A} \rightarrow [0,1]$.
\end{definition}
The base score function in QBAFs assigns an apriori belief to arguments.
QBAFs can be represented graphically (as in Figure~\ref{fig_tdqbaf}) using nodes to represent arguments and edges to show the relations between them. Then QBAFs are said to be \emph{(a)cyclic} if the graphs representing them are (a)cyclic.

In this paper, we use the QE gradual semantics~\cite{Potyka18} to evaluate the \emph{strength} of arguments in QBAFs. Like most QBAF semantics, it computes strength values 
iteratively by initializing the strength value of each argument with its base score and repeatedly applying an update function.
Let us represent the strength of arguments in the $i$-th iteration by a function 
$$\sigma^i: \mathcal{A} \rightarrow [0,1],$$ 
where
$\sigma^0(\alpha) = \tau(\alpha)$ for all $\alpha \in \mathcal{A}$.
In order to compute $\sigma^{i+1}$ from $\sigma^i$, the update function first computes the \emph{energy} $E^i_\alpha$ of attackers and supporters of each argument $\alpha$ defined by 
$$E^i_\alpha=\sum_{\left \{ \beta \in \mathcal{A} \mid (\beta,\alpha) \in \mathcal{R^{+}} \right \} }\sigma^i(\beta) - \sum_{\left \{ \beta \in \mathcal{A} \mid (\beta,\alpha) \in \mathcal{R^{-}} \right \} }\sigma^i(\beta).$$
It then computes the strength in the next iteration via
$$
\sigma^{i+1}(\alpha)= 
    \begin{cases}
        \tau(\alpha)-\tau(\alpha)\cdot\frac{(E^i_\alpha)^2}{1+(E^i_\alpha)^2} & if\ E^i_\alpha \leq 0;\\
        \tau(\alpha)+(1-\tau(\alpha))\cdot\frac{(E^i_\alpha)^2}{1+(E^i_\alpha)^2} & if\ E^i_\alpha > 0.\\
    \end{cases}
$$
The final dialectical strength of each argument $\alpha$ is then defined as the limit $\lim_{t \rightarrow \infty} \sigma^{t}(\alpha)$.
In cyclic graphs, the strength values may start oscillating and the limit may not exist \cite{mossakowski2018modular}.
In all known cases, the problem can be solved by \emph{continuizing} the semantics \cite{Potyka18,PotykaBooth24}. However, we do not have
space to discuss these issues in more detail here and will just restrict to examples where the strength values converge.




To better understand the QE gradual semantics, let us look at an example.
\begin{example}
\label{example_QE}

Consider the QBAF in Figure \ref{fig_qe_semantics}, where the base scores are given as $\tau(\alpha)=0.8$, $\tau(\beta)=0.6$, $\tau(\gamma)=0.9$, and $\tau(\delta)=0.7$.
\begin{figure}[t]
	\centering
		\includegraphics[width=0.3\columnwidth]{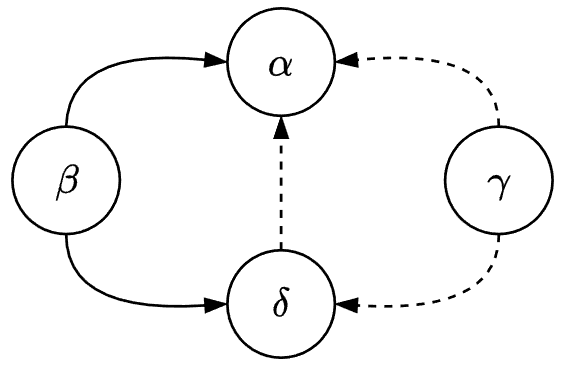}
	\caption{Example of a QBAF structure for computing the QE gradual semantics.}
	\label{fig_qe_semantics}
\end{figure}
Since $\beta$ and $\gamma$ have no parents, we have $E^i_\beta = E^i_\gamma = 0$ for all $i$ and thus
$\sigma(\beta) = \tau(\beta) = 0.6$ and $\sigma(\gamma) = \tau(\gamma) = 0.9$.
For $\delta$, we have $E^i_\delta = \sigma^i(\gamma) - \sigma^i(\beta) = 0.3$ for all $i$, hence $\sigma(\delta) = \tau(\delta) + (1 - \tau(\delta)) \cdot 0.3^2/(1 + 0.3^2) = 0.72$.
For $\alpha$, we have $E^i_\alpha = \sigma^i(\gamma) + \sigma^i(\delta) - \sigma^i(\beta) = 1.02$ for all $i\geq 1$. Hence, $\sigma(\alpha) = \tau(\alpha) + (1 - \tau(\alpha)) \cdot 1.02^2/(1 + 1.02^2) = 0.90$.
\end{example}

In the remainder, unless specified otherwise, we assume as given a generic QBAF 
$\mathcal{Q}=\langle \mathcal{A}, \mathcal{R}^{-}, \mathcal{R}^{+}, \tau \rangle$
and we let $\mathcal{R}=\mathcal{R}^{-}  \cup  \mathcal{R}^{+}$ 
We will often need to restrict QBAFs to a subset of the arguments or edges, or change the base score function, as follows. 

\begin{notation}
For ${\mathcal{U}} \subseteq  \mathcal{A}$, let $\mathcal{Q}^{|_{\mathcal{U}}} =\langle \mathcal{A}\cap \mathcal{U}, \mathcal{R}^{-}, \mathcal{R}^{+}, \tau \rangle$. Then, for any $\alpha\in \mathcal{A}$, we let $\sigma_{\mathcal{U}}(\alpha)$ denote the strength of $\alpha$ in $\mathcal{Q}^{|_{\mathcal{U}}}$.
\end{notation}

\begin{notation}
For ${\mathcal{S}} \subseteq  \mathcal{R}$, let $\mathcal{Q}^{|_{\mathcal{S}}} =\langle \mathcal{A}, \mathcal{R}^{-}\cap \mathcal{S}, \mathcal{R}^{+} \cap \mathcal{S}, \tau \rangle$. Then, for any $\alpha\in \mathcal{A}$, we let $\sigma_{\mathcal{S}}(\alpha)$ denote the strength of $\alpha$ in $\mathcal{Q}^{|_{\mathcal{S}}}$.
\end{notation}

\begin{notation}
For $\tau':\mathcal{A} \rightarrow [0,1]$ a base score function, let $\mathcal{Q}^{|_{\tau'}} =  \langle \mathcal{A}, \mathcal{R}^{-}, \mathcal{R}^{+}, \tau' \rangle$. Then, for any $\alpha\in \mathcal{A}$, we let $\sigma_{\tau'}(\alpha)$ denote the strength of $\alpha$ in $\mathcal{Q}^{|_{\tau'}}$.
\end{notation}

\subsection{Truth Discovery QBAFs (TD-QBAFs)}

TD-QBAFs allow reasoning about
truth discovery problems using quantitative 
argumentation.
Truth discovery problems can be described
concisely as \emph{truth discovery networks (TDNs)} \cite{singleton2022towards}.
Formally, a TDN is a tuple 
$\tdn = (\sources, \objects, \domain, \reports)$ consisting of 
a finite set of \emph{sources} $\sources$, 
a finite set of \emph{objects} $\objects$,
a set $\domain = \{D_o\}_{o \in O}$ of \emph{domains} of the objects,
and a set of \emph{reports} $\reports \subseteq \sources \times \objects \times V$, where $V = \bigcup_{o \in O}{D_o}$, and for all
$(s, o, v) \in \reports$, we have $v \in D_o$, and there is no $(s, o, v') \in \reports$ with $v \neq v'$.
Given a TDN $\tdn$, we are interested in a
truth discovery operator that assigns a
trust score to each source and
each claim \cite{singleton2022towards}.

Singleton suggested to reason about TDNs using bipolar argumentation frameworks, where we have bi-directional support edges between sources and their claims (trustworthy sources make claims more believable, and, conversely, believable claims make sources more trustworthy) and
contradictory claims attack each other
\cite{singleton2020link}.
TD-QBAFs implement this idea with QBAFs, where
sources have a base score of $0.5$ (we are initially indifferent about the trustworthiness of sources) and claims have a base score of $0$
(we do not believe anything without evidence).
\begin{definition}[TD-QBAF induced from a TDN]
The TD-QBAF 
induced from the TDN $\tdn = (\sources, \objects, \domain, \reports)$
is defined as $Q = (\args, \attacker, \supporter, \baseScore)$,
where 
    $\args = \sources \cup \{(o,v) \mid \exists s \in \sources: (s,o,v) \in \reports\}$,
    $\attacker = \{(c,c') \in \args^2 \cap C^2 \mid \obj(c) = \obj(c'), \val(c) \neq \val(c')\}$,
    $\supporter =  \{(s,(o,v)), ((o,v),s) \mid (s,o,v) \in \reports  \}$.
    $\baseScore(s) = 0.5$ for all $s \in \sources$ and $\baseScore(c) = 0$ for all $c \in C$.
\end{definition}
Every QBAF semantics gives rise to a truth discovery operator that is defined by associating
each source and claim with its final strength under the semantics. The semantical properties
of QBAF semantics like balance and monotonicity
directly translate to meaningful guarantees
for the derived trust scores.

\subsection{Argument Attribution Explanations}
In order to explain trust scores in TD-QBAFs, we recall the removal-based and Shapley-based AAEs. AAEs aim at evaluating the impact of an argument on a given topic argument.
The removal-based AAEs proposed by~\cite{delobelle2019interpretability} measure how the strength of the topic argument changes if an argument is removed. 
\begin{definition}[Removal-based AAEs]
\label{removal_AAE}
Let $\alpha,\beta \in \mathcal{A}$. 
The \emph{removal-based AAE from $\beta$ to $\alpha$ under $\sigma$} is:
$$
\varphi_{\sigma}^{\alpha}(\beta)=\sigma(\alpha)-\sigma_{\mathcal{A} \setminus \{\beta\}}(\alpha).
$$
\end{definition}

The Shapley-based AAEs~\cite{vcyras2022dispute,kampik2024contribution} use the Shapley value from coalitional game theory \cite{shapley1951notes} to assign attributions.
Each argument in a QBAF is seen as a \emph{player} that can contribute to the strength of the topic argument. Intuitively, Shapley-based AAEs look at all possible ways how
the argument could be added to the QBAF and average its impact on the topic argument.
\begin{definition}[Shapley-based AAEs]
\label{shapley_AAE}
Let $\alpha,\beta \in \mathcal{A}$. 
The \emph{Shapley-based AAE from $\beta$ to $\alpha$ under $\sigma$} is:
$$
\psi_{\sigma}^{\alpha}(\beta)=\sum_{\mathcal{U} \subseteq \mathcal{A} \setminus \{\alpha,\beta\}} \frac{(\left| \mathcal{A} \setminus \{\alpha\}  \right| - \left| \mathcal{U} \right| -1)!\left| \mathcal{U} \right|!}{\left| \mathcal{A} \setminus \{\alpha\} \right|!} \left[ \sigma_{\mathcal{U} \cup \{\beta\}}(\alpha)-\sigma_{\mathcal{U}}(\alpha)\right].
$$
\end{definition}

\subsection{Relation Attribution Explanations}

RAEs are similar to AAEs, but measure the impact of edges rather than the impact of arguments. 
Analogous to the idea of removal-based AAEs~\cite{delobelle2019interpretability}, we consider the removal-based RAEs.
\begin{definition}[Removal-based RAEs]
\label{removal_RAE}
Let $\alpha \in \mathcal{A}$ and $r \in \mathcal{R}$. 
The \emph{removal-based RAE from $r$ to $\alpha$ under $\sigma$} is:
$$
\lambda_{\sigma}^{\alpha}(r)=\sigma(\alpha)-\sigma_{\mathcal{R} \setminus \{r\}}(\alpha).
$$
\end{definition}

Shapley-based RAEs~\cite{amgoud2017measuring,YIN_RAE_IJCAI} share the same idea with Shapley-based AAEs, but the attribution objects are changed from arguments to edges.
\begin{definition}[Shapley-based RAEs]
\label{shap_RAE}
Let $\alpha \in \mathcal{A}$ and $r \in \mathcal{R}$. 
The \emph{Shapley-based RAE from $r$ to $\alpha$ under $\sigma$} is:
$$
\phi_{\sigma}^{\alpha}(r)=\sum_{\mathcal{S} \subseteq \mathcal{R} \setminus \{r\}} \frac{(\left| \mathcal{R} \right| - \left| \mathcal{S} \right| -1)!\left| \mathcal{S} \right|!}{\left| \mathcal{R} \right|!} \left[ \sigma_{\mathcal{S} \cup \{r\}}(\alpha)-\sigma_{\mathcal{S}}(\alpha)\right].
$$
\end{definition}

\section{Explaining TD-QBAFs with AAEs and RAEs}

\subsection{Settings}
To compare the different AEs, 
we explain the strength of argument $c5$ in Figure \ref{fig_tdqbaf}.
Since there are $17$ arguments and $32$ edges in Figure~\ref{fig_tdqbaf}, computing Shapley-based AAEs and RAEs exactly is prohibitively expensive. 
We therefore apply the approximation algorithm from~\cite{YIN_RAE_IJCAI} that approximates the Shapley values using sampling (we set the sample size to $1000$). 

We report the removal and Shapley-based AAEs and RAEs in Figure \ref{fig_AAE_2} and \ref{fig_RAE_2} \footnote{The numerical AAEs and RAEs can be found in the Appendix}. In addition, to provide intuitive explanations for argument $c5$, we visualize the removal and Shapley-based AAEs and RAEs as shown in Figure \ref{fig_AAE_2} and \ref{fig_RAE_2}, where blue/red arguments or edges denote positive/negative AAEs or RAEs. The darkness of the color of arguments and the thickness of the edges denote the magnitude of the their AAEs and RAEs, respectively\footnote{The code of all experiments is available at \url{https://github.com/XiangYin2021/TD-QBAF-AAE-RAE}.}.

\subsection{Results and Analysis for AAEs}

\begin{figure}[t]
    \centering
    \includegraphics[width=1.0\columnwidth]{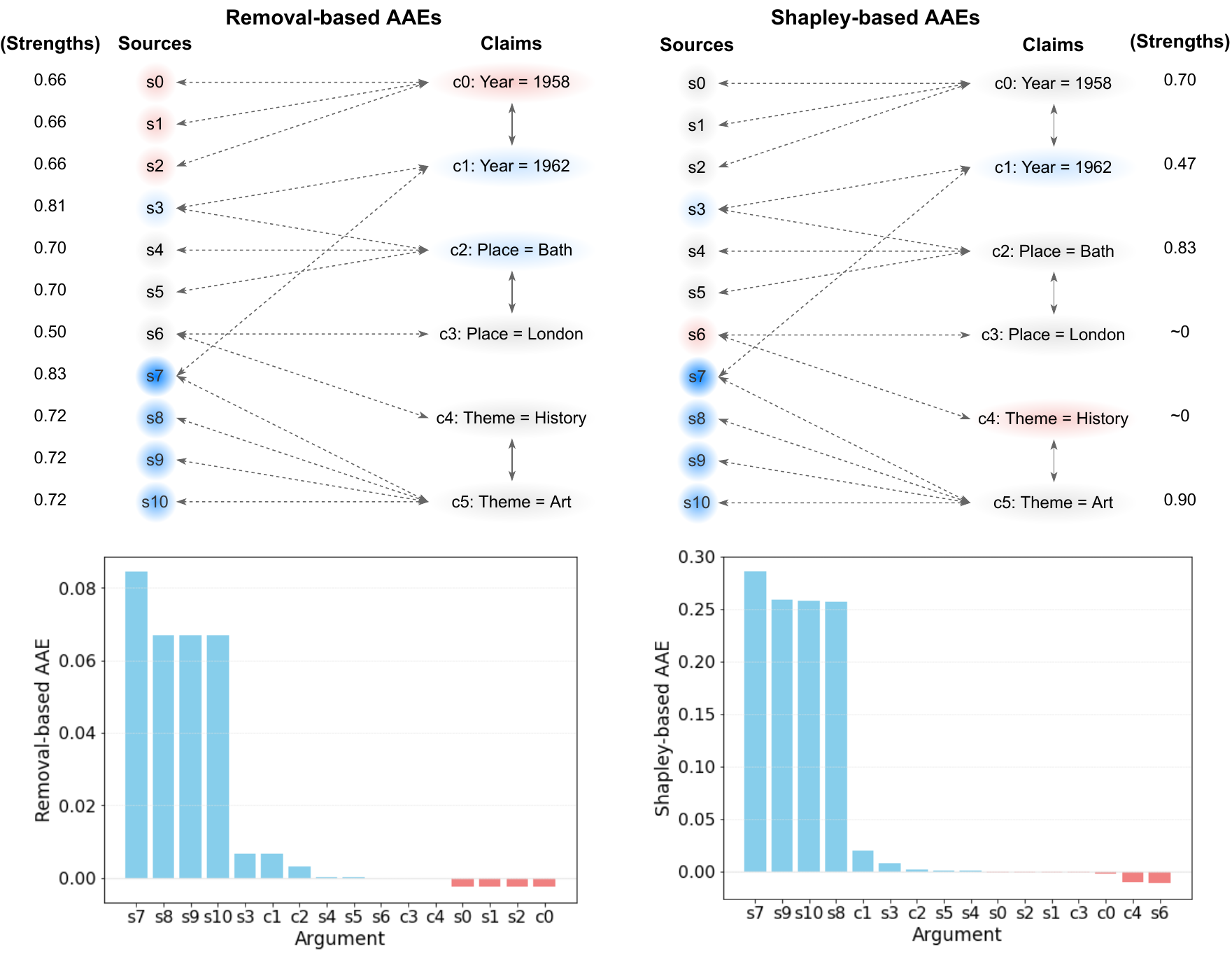}
    \caption{Removal and Shapley-based AAEs for the topic argument $c5$ of TD-QBAF in Figure \ref{fig_tdqbaf}. (Blue/red/grey nodes denote positive/negative/negligible AAEs, respectively. The darkness of nodes represents the magnitude of their AAE values.)}
    \label{fig_AAE_2}
\end{figure}

Figure \ref{fig_AAE_2} shows the results of removal and Shapley-based AAEs.


For the \textbf{removal-based AAEs}, we observe that $s7$, $s8$, $s9$, and $s10$ have noticeably positive influences on $c5$, followed by minor positive influences from $s3$, $c1$, and $c2$. This is because $s7$ to $s10$ are direct supporters for $c5$, whereas $s3$, $c1$, and $c2$ indirectly support $c5$. Specifically, $c2$ supports $s3$, $s3$ supports $c1$, $c1$ supports $s7$, and then $s7$ supports $c5$, meaning $s3$, $c1$, and $c2$ all indirectly support $c5$. These indirect influences also explain why the AAEs of $s3$, $c1$, and $c2$ are much smaller than those of $s7$ to $s10$. 
Besides, since $s7$ is supported by $c1$, its AAE is slightly larger than those of $s8$ to $s10$, which have consistent AAEs due to their symmetrical structure to $c5$.
In contrast, $s0$, $s1$, $s2$, and $c0$ have minor negative influences on $c5$ because $c0$ attacks $c1$, an indirect supporter for $c5$. Furthermore, $s0$ to $s2$ support $c0$, and thus they have negative influences on $c5$ as well. However, their negative influences are not obvious due to the indirect influences.
Finally, the remaining arguments have AAEs close to $0$, indicating their negligible influences on $c5$.

When considering the \textbf{Shapley-based AAEs}, the results are similar to those of removal-based AAEs, where $s7$ to $s10$ still have significant influences on $c5$. Unlike removal-based AAEs, however, we notice that both $c4$ and $s6$ have minor negative influences on $c5$. This is because $c4$ directly attacks $c5$, while $s6$ indirectly attacks $c5$ by supporting $c4$, although the QE strength of $c4$ is very small (close to $0$). Also, the negative influences of $s0$ to $s2$ and $c0$ and positive influence of $c2$ are relatively negligible compared with those of in removal-based AAEs due to their indirect connection to $c5$.

In this case study, both removal and Shapley-based AAEs can effectively capture the main influential arguments despite having some tiny differences in those low contributing arguments. 
This is mainly because of their different mechanisms of computing the AAEs. Another important reason is probably due to the approximation algorithm used for Shapley-based AAEs, leading to different AAEs even with the same sample size for the coalitions.
We also noticed that the qualitative influence (the sign) of those Shapley-based AAEs close to $0$ is sensitive when applying the approximation algorithm, thus we do not visualize those close to $0$. However, this should not be a concern since their influence is negligible.

\subsection{Results and Analysis for RAEs}

\begin{figure}[t]
    \centering
    \includegraphics[width=1.0\columnwidth]{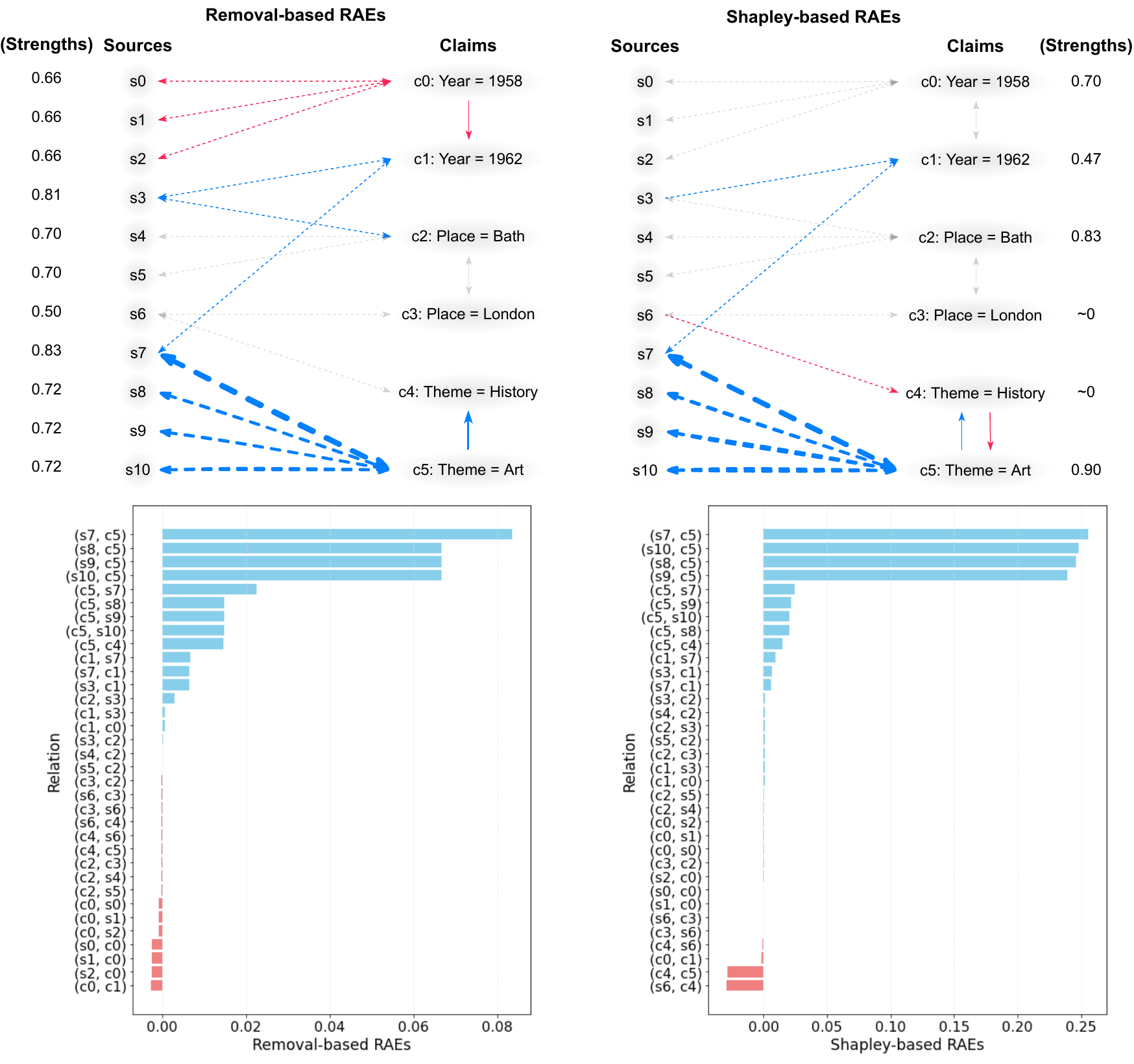}
    \caption{Removal and Shapley-based RAEs for the topic argument $c5$ of TD-QBAF in Figure \ref{fig_tdqbaf}. (Blue/red/grey edges denote positive/negative/negligible RAEs, respectively. The darkness of edges represents the magnitude of their RAE values.)}
    \label{fig_RAE_2}
\end{figure}


Figure \ref{fig_RAE_2} shows the results of removal and Shapley-based RAEs.

Let us first discuss the \textbf{removal-based RAEs}.
We see that $(s7,c5)$ has the largest positive impact on $c5$. Following closely are $(s8,c5)$, $(s9,c5)$, and $(s10,c5)$, which also have notably positive influences on $c5$ because they are direct incoming supports to $c5$.
There are also four outgoing supports from $c5$, namely $(c5,s7)$, $(c5,s8)$, $(c5,s9)$, and $(c5,s10)$, with positive influences but their RAEs are greatly smaller than that of the previous four as they are indirect supports. For instance, $c5$ first supports $s7$, and then $s7$ supports $c5$, indicating the indirect positive influence of $(c5,s7)$.
Additionally, $(c5,c4)$ also contributes positively to $c5$ because $c5$ attacks its attacker $c4$, thus the attack from $c4$ to $c5$ is weakened.
We can also observe some marginal influences, such as the positive influences provided by $(c1,s7)$, $(s3,c1)$, and $(c2,s3)$ on $c5$, while the negative influences from $(c0,c1)$, $(s0,c0)$, $(s1,c0)$, and $(s2,c0)$.
The remaining edges have RAEs close to $0$, showing their negligible influence on $c5$.

When it comes to the \textbf{Shapley-based RAEs}, which have similar effects to removal-based RAEs, the four incoming supports to $c5$ are still the major contributors, and the four outgoing supports from $c5$ have minor RAEs. Different from removal-based RAEs, Shapley-based RAEs capture some different negligible influences, such as the negative influence by $(c4,c5)$ and $(s6,c4)$. However, Shapley-based RAEs also disregard some tiny influences, like $(s0,c0)$, $(s1,c0)$, and $(s2,c0)$, which are shown by removal-based RAEs.

In this case study, both removal and Shapley-based RAEs have a consistent ranking for the main influential edges despite having some tiny differences in those low contributing edges. The reasons are the same as we discussed above.

Let us further compare the results of AAEs and RAEs.
In this case study, we observe some connections between AAEs and RAEs.
For example, in both AAEs, the top-$4$ influential arguments are $s7$ to $s10$, while in both RAEs, the outgoing edges from these arguments ($(s7,c5)$, $(s8,c5)$, $(s9,c5)$, and $(s10,c5)$) also rank in the top-$4$.
In addition, $s0$ to $s4$ and $c0$ to $c2$ have minor influences in the removal-based AAEs, while their incoming or outgoing edges also have minor influences in the removal-based RAEs. A similar phenomenon can be found in the Shapley-based AAEs and RAEs.
While it is expected that the RAEs for outgoing edges of important arguments are relatively high, the consistency observed across different sets of arguments and edges is noteworthy. Besides, we found that removal or Shapley-based AAE of an argument does not necessarily equate to the sum of RAEs of all its incoming and outgoing edges, which goes against a reasonable expectation. We will leave the investigation of their formal relationships for future work.

\section{Conclusion}
Since most existing applications of AAEs and RAEs focus on acyclic QBAFs, this paper investigated their applicability in cyclic QBAFs.
First, we found that AAEs and RAEs can provide intuitive explanations.
By displaying the ranking of arguments or edges, it is easy to identify the most influential arguments or edges in the QBAF without delving into the complex (cyclic) structure of the QBAFs, particularly in TD-QBAFs where the number of arguments is typically large and the connections between source arguments and claim arguments are bi-directional.
Second, AAEs and RAEs can provide interesting or even surprising explanations.
For example, in the case study provided earlier, one might overlook the influence between claim arguments $c1$ and $c5$ because they are in different topics (\emph{Year=1962} and \emph{Theme=Art}), but AEs demonstrate that $c1$ can contribute to $c5$ through $s7$.
Third, RAEs provide more fine-grained explanations than AAEs. 
This is because when computing AAEs, such as removal-based AAEs, removing an argument means removing all the incoming and outgoing edges associated with that argument, whereas RAEs offer a more detailed insight by processing every incoming and outgoing edge individually.
One can choose between them depending on the granularity for practical use.

For future work, it would be worthwhile to investigate how different gradual semantics influence AAEs and RAEs, 
because the property satisfaction of semantics have an influence on the property satisfaction of explanations.
Additionally, the formal relationship between AAEs and RAEs requires further exploration. However, we believe AAEs and RAEs can complement each other, providing a deeper and more comprehensive understanding of the internal mechanisms of QBAFs, particularly the interactions between arguments and edges in complex QBAFs.

\begin{acknowledgments}
This research was partially funded by the  European Research Council (ERC) under the
European Union’s Horizon 2020 research and innovation programme (grant
agreement No. 101020934, ADIX) and by J.P. Morgan and by the Royal
Academy of Engineering under the Research Chairs and Senior Research
Fellowships scheme. Any views or opinions expressed herein are solely those of the authors.
\end{acknowledgments}

\bibliography{sample}

\appendix

\section*{Additional Results for AAEs and RAEs}

\begin{table}[h]
    \centering
    \small
    \caption{Comparison of removal-based AAEs and Shapley-based AAEs (in descending order) for the argument $c5$ of TD-QBAF in Figure \ref{fig_tdqbaf}. Note that they are in different scales.}
    \begin{tabular}{cc|cc}
        \hline
        \textbf{Argument} & \textbf{Removal-based AAE} & \textbf{Argument} & \textbf{Shapley-based AAE} \\
        \hline
        s7  & 0.084304029 & s7  & 0.285373360 \\
        s8  & 0.066738248 & s9  & 0.259206533 \\
        s9  & 0.066738248 & s10 & 0.257762474 \\
        s10 & 0.066738248 & s8  & 0.256392126 \\
        s3  & 0.006673061 & c1  & 0.020544840 \\
        c1  & 0.006635552 & s3  & 0.007852405 \\
        c2  & 0.002938110 & c2  & 0.001789997 \\
        s4  & 0.000076913 & s5  & 0.000810191 \\
        s5  & 0.000076913 & s4  & 0.000789093 \\
        s6  & -0.000008421 & s0 & -0.000593859 \\
        c3  & -0.000008421 & s2 & -0.000917430 \\
        c4  & -0.000008423 & s1 & -0.001164825 \\
        s0  & -0.002444482 & c3 & -0.001309005 \\
        s1  & -0.002444482 & c0 & -0.001711158 \\
        s2  & -0.002444482 & c4 & -0.010154039 \\
        c0  & -0.002476209 & s6 & -0.010892383 \\
        \hline
    \end{tabular}
\end{table}

\begin{table}[t]
    \centering
    \small
    \caption{Comparison of removal-based RAEs and Shapley-based RAEs (in descending order) for the argument $c5$ of TD-QBAF in Figure \ref{fig_tdqbaf}. Note that they are in different scales.}
    \begin{tabular}{cc|cc}
        \hline
        \textbf{Relation} & \textbf{Removal-based RAE} & \textbf{Relation} & \textbf{Shapley-based RAE} \\
        \hline
        (s7, c5)  & 0.083473613 & (s7, c5)  & 0.255513421 \\
        (s8, c5)  & 0.066745475 & (s10, c5) & 0.247761961 \\
        (s9, c5)  & 0.066745475 & (s8, c5)  & 0.245927825 \\
        (s10, c5) & 0.066745475 & (s9, c5)  & 0.238930772 \\
        (c5, s7)  & 0.022507211 & (c5, s7)  & 0.024524066 \\
        (c5, s8)  & 0.014968725 & (c5, s9)  & 0.022019375 \\
        (c5, s9)  & 0.014968725 & (c5, s10) & 0.020724751 \\
        (c5, s10) & 0.014968725 & (c5, s8)  & 0.020059231 \\
        (c5, c4)  & 0.014703252 & (c5, c4)  & 0.015153831 \\
        (c1, s7)  & 0.006793938 & (c1, s7)  & 0.009541657 \\
        (s7, c1)  & 0.006577898 & (s3, c1)  & 0.006460594 \\
        (s3, c1)  & 0.006576020 & (s7, c1)  & 0.005974419 \\
        (c2, s3)  & 0.002946488 & (s3, c2)  & 0.001282426 \\
        (c1, s3)  & 0.000805779 & (s4, c2)  & 0.001206074 \\
        (c1, c0)  & 0.000695966 & (c2, s3)  & 0.001140980 \\
        (s3, c2)  & 0.000212036 & (s5, c2)  & 0.001083422 \\
        (s4, c2)  & 0.000085191 & (c2, c3)  & 0.000928671 \\
        (s5, c2)  & 0.000085191 & (c1, s3)  & 0.000881815 \\
        (c3, c2)  & -0.000007913 & (c1, c0)  & 0.000834337 \\
        (s6, c3)  & -0.000007913 & (c2, s5)  & 0.000670102 \\
        (c3, s6)  & -0.000007913 & (c2, s4)  & 0.000631869 \\
        (s6, c4)  & -0.000007913 & (c0, s2)  & 0.000586827 \\
        (c4, s6)  & -0.000007913 & (c0, s1)  & 0.000558687 \\
        (c4, c5)  & -0.000007915 & (c0, s0)  & 0.000558250 \\
        (c2, c3)  & -0.000115977 & (c3, c2)  & 0.000164534 \\
        (c2, s4)  & -0.000120350 & (s2, c0)  & 0.000144884 \\
        (c2, s5)  & -0.000120350 & (s0, c0)  & -0.000017152 \\
        (c0, s0)  & -0.000641070 & (s1, c0)  & -0.000036753 \\
        (c0, s1)  & -0.000641070 & (s6, c3)  & -0.000276491 \\
        (c0, s2)  & -0.000641070 & (c3, s6)  & -0.000465298 \\
        (s0, c0)  & -0.002436075 & (c4, s6)  & -0.001147533 \\
        (s1, c0)  & -0.002436075 & (c0, c1)  & -0.001659705 \\
        (s2, c0)  & -0.002436075 & (c4, c5)  & -0.028383722 \\
        (c0, c1)  & -0.002598473 & (s6, c4)  & -0.029099303 \\
        \hline
    \end{tabular}
\end{table}

\end{document}